\documentclass[11pt,a4paper]{article}
\usepackage[acceptedWithA]{tacl2018v2}
\usepackage{times}
\usepackage{latexsym}
\usepackage{graphicx}
\usepackage{amsmath}
\usepackage{amssymb}
\usepackage{multirow}
\usepackage{url}
\usepackage{color}
\usepackage[utf8]{inputenc}

\title{Semantic Neural Machine Translation using AMR}

\author{Linfeng Song$^1$, Daniel Gildea$^1$, Yue Zhang$^2$, Zhiguo Wang$^3$ \and Jinsong Su$^4$ \\
  $^1$Department of Computer Science, University of Rochester, Rochester, NY 14627 \\
  $^2$School of Engineering, Westlake University, China \\
  $^3$IBM T.J. Watson Research Center, Yorktown Heights, NY 10598 \\
  $^4$Xiamen University, Xiamen, China \\
  {\tt $^1$\{lsong10,gildea\}@cs.rochester.edu~~$^2$yue.zhang@wias.org.cn} \\
  {\tt ~~~~~$^3$zgw.tomorrow@gmail.com~~$^4$jssu@xmu.edu.cn} \\}

\date{}

\begin{document}
\maketitle
\begin{abstract}
  It is intuitive that semantic representations can be useful for machine translation, mainly because they can help in enforcing meaning preservation and handling data sparsity (many sentences correspond to one meaning) of machine translation models.
  On the other hand, little work has been done on leveraging semantics for neural machine translation (NMT).
  In this work, we study the usefulness of AMR (short for abstract meaning representation) on NMT.
  Experiments on a standard English-to-German dataset show that incorporating AMR as additional knowledge can significantly improve a strong attention-based sequence-to-sequence neural translation model. 
\end{abstract}

\section{Introduction}

It is intuitive that semantic representations ought to
be relevant to machine translation, given that the
task is to produce a target language sentence with the
same meaning as the source language input.
Semantic representations formed the core of the earliest symbolic
machine translation systems, and have been applied
to statistical but non-neural systems as well.

\begin{figure}
\centering
\includegraphics[width=0.9\linewidth]{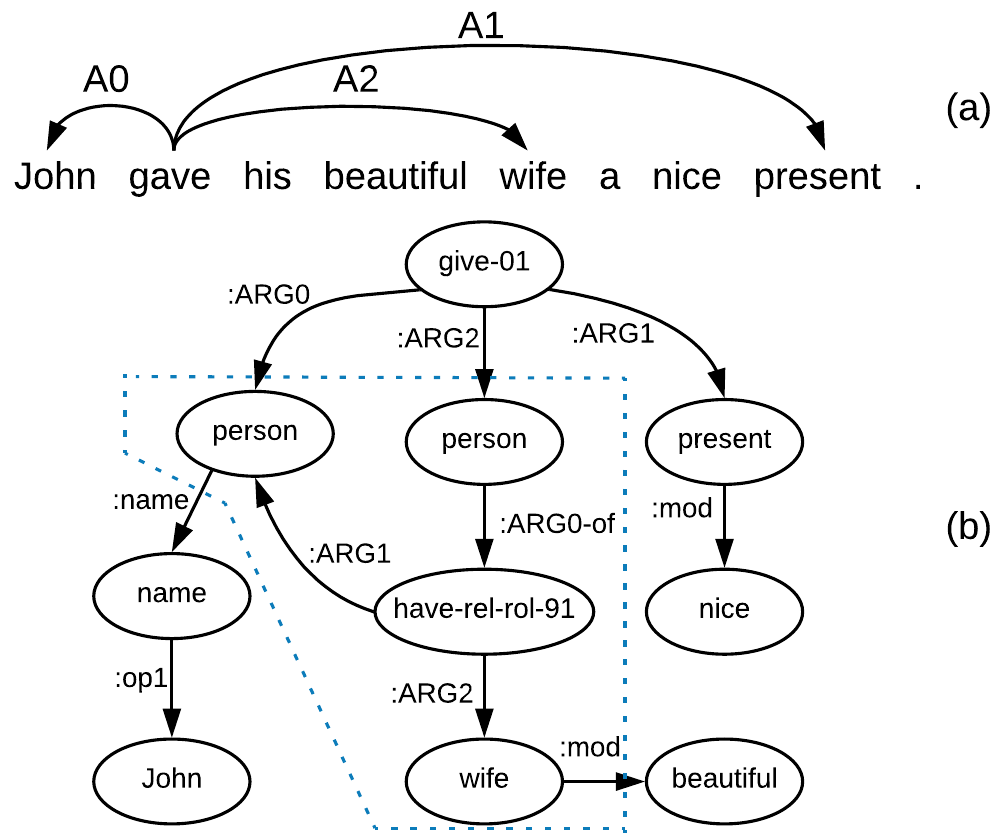}
\caption{(a) A sentence with semantic roles annotations, (b) the corresponding AMR graph of that sentence.}
\label{fig:example}
\vspace{-0.5em}
\end{figure}

Leveraging syntax for neural machine translation (NMT) has been an active research topic \cite{stahlberg-EtAl:2016:P16-2,aharoni-goldberg:2017:Short,li-EtAl:2017:Long,chen-EtAl:2017:Long6,bastings-EtAl:2017:EMNLP2017,wu2017improved,chen2017syntax}.
On the other hand, exploring semantics for NMT has so far received relatively little attention.
Recently, \newcite{diego-EtAl:2018:PAPERS} exploited semantic role labeling (SRL) for NMT, showing that the predicate-argument information from SRL can improve the performance of an attention-based sequence-to-sequence model by alleviating the ``argument switching'' problem,\footnote{flipping arguments corresponding to different roles} one frequent and severe issue faced by NMT systems \cite{isabelle-cherry-foster:2017:EMNLP2017}.
Figure \ref{fig:example} (a) shows one example of semantic role information, which only captures the relations between a predicate ({\em gave}) and its arguments ({\em John}, {\em wife} and {\em present}).
Other important information, such as the relation between {\em John} and {\em wife}, can not be incorporated.

In this paper, we explore the usefulness of abstract meaning representation (AMR) \cite{banarescu-EtAl:2013:LAW7-ID} as a semantic representation for NMT\@.
AMR is a semantic formalism that encodes the meaning of a sentence as a rooted, directed graph.
Figure \ref{fig:example} (b) shows an AMR graph, in which the nodes (such as {\em give-01} and {\em John}) represent the concepts,
and edges (such as {\em :ARG0} and {\em :ARG1}) represent the relations between concepts they connect.
Comparing with semantic roles, AMRs capture more relations, such as the relation between {\em John} and {\em wife} (represented by the subgraph within dotted lines).
In addition, AMRs directly capture entity relations and abstract away inflections and function words.
As a result, they can serve as a source of knowledge for machine translation that is orthogonal to the textual input.
Furthermore, structural information from AMR graphs can help reduce data sparsity, when training data is not sufficient for large-scale training.

Recent advances in AMR parsing keep pushing the boundary of state-of-the-art performance \cite{flanigan-EtAl:2014:P14-1,artzi-lee-zettlemoyer:2015:EMNLP,pust-EtAl:2015:EMNLP,K15-1004,flanigan-EtAl:2016:SemEval,buys-blunsom:2017:Long,konstas-EtAl:2017:Long,wang-xue:2017:EMNLP2017,P18-1037,peng:2018:ACL,P18-1170,D18-1198}, and have made it possible for automatically-generated AMRs to benefit down-stream tasks, such as question answering \cite{mitra2015addressing}, summarization \cite{takase-EtAl:2016:EMNLP2016}, and event detection \cite{li-EtAl:2015:CNewsStory}.
However, to our knowledge, no existing work has exploited AMR for enhancing NMT.

We fill in this gap, taking an attention-based sequence-to-sequence system as our baseline, which is similar to \newcite{bahdanau2015neural}.
To leverage knowledge within an AMR graph, we adopt a graph recurrent network (GRN) \cite{song-EtAl:acl2018,zhang-EtAl:acl2018} as the AMR encoder.
In particular, a full AMR graph is considered as a single state, with nodes in the graph being its substates. 
State transitions are performed on the graph recurrently, allowing substates to exchange information through edges. 
At each recurrent step, each node advances its current state by receiving information from the current states of its adjacent nodes. 
Thus, with increasing numbers of recurrent steps, each word receives information from a larger context. Figure~\ref{fig:encoder} shows the recurrent transition, where each node works simultaneously.
Compared with other methods for encoding AMRs \cite{konstas-EtAl:2017:Long}, GRN keeps the original graph structure,
and thus no information is lost \cite{song-EtAl:acl2018}.
For the decoding stage, two separate attention mechanisms are adopted in the AMR encoder and sequential encoder, respectively.

Experiments on WMT16 English-German data (4.17M) show that adopting AMR significantly improves a strong attention-based sequence-to-sequence baseline (25.5 vs 23.7 BLEU).
When trained with small-scale (226K) data, the improvement increases (19.2 vs 16.0 BLEU), which shows that the structural information from AMR can alleviate data sparsity when training data are not sufficient.
To our knowledge, we are the first to investigate AMR for NMT.

Our code and parallel data (training/dev/test) with automatically parsed AMRs are available at https://github.com/freesunshine0316/semantic-nmt.

\section{Related work}

Most previous work on exploring semantics for statistical machine translation (SMT) studies the usefulness of predicate-argument structure from semantic role labeling \cite{wong-mooney:2006:HLT-NAACL06-Main,wu-fung:2009:NAACLHLT09-Short,liu-gildea:2010:PAPERS,baker2012modality}.
\newcite{jones-EtAl:2012:PAPERS} first convert Prolog expressions into graphical meaning representations, leveraging synchronous hyperedge replacement grammar to parse the input graphs while generating the outputs.
Their graphical meaning representation is different from AMR under a strict definition, and their experimental data are limited to 880 sentences.
We are the first to investigate AMR on a large-scale machine translation task.

Recently, \newcite{diego-EtAl:2018:PAPERS} investigate semantic role labeling (SRL) on neural machine translation (NMT).
The predicate-argument structures are encoded via graph convolutional network (GCN)  layers \cite{kipf2017semi}, which are laid on top of regular BiRNN or CNN layers. 
Our work is in line with exploring semantic information, but different in exploiting AMR rather than SRL for NMT\@.
In addition, we leverage a graph recurrent network (GRN) \cite{song-EtAl:acl2018,zhang-EtAl:acl2018} for modeling AMRs rather than GCN, which is formally consistent with the RNN sentence encoder.
Since there is no one-to-one correspondence between AMR nodes and source words, we adopt a doubly-attentive LSTM decoder, which is another major difference from \newcite{diego-EtAl:2018:PAPERS}.

GRNs have recently been used to model graph structures in NLP tasks.
In particular, \newcite{zhang-EtAl:acl2018} use a GRN model to represent raw sentences by building a graph structure of neighboring words and a sentence-level node, showing that the encoder outperforms BiLSTMs and Transformer \cite{NIPS2017_7181} on classification and sequence labeling tasks; 
\newcite{song-EtAl:acl2018} build a GRN for encoding AMR graphs for text generation, showing that the representation is superior compared to BiLSTM on serialized AMR. 
We extend \newcite{song-EtAl:acl2018} by investigating the usefulness of AMR for neural machine translation.
To our knowledge, we are the first to use GRN for machine translation.

In addition to GRNs and GCNs, there have been other graph neural networks, such as graph gated neural network (GGNN) \cite{li2015gated,P18-1026}. 
Since our main concern is to empirically investigate the effectiveness of AMR for NMT, we leave it to future work to compare GCN, GGNN, and GRN for our task.

\section{Baseline: attention-based BiLSTM}
\label{sec:base}

We take the attention-based sequence-to-sequence model of \newcite{bahdanau2015neural} as the baseline, but use LSTM cells \cite{hochreiter1997long} instead of GRU cells \cite{cho-EtAl:2014:EMNLP2014}.

\subsection{BiLSTM encoder}
\label{sec:base_enc}

The encoder is a bi-directional LSTM on the source side.
Given a sentence, two sequences of states $[\overleftarrow{\boldsymbol{h}}_1, \overleftarrow{\boldsymbol{h}}_2, \dots, \overleftarrow{\boldsymbol{h}}_N]$ and $[\overrightarrow{\boldsymbol{h}}_1, \overrightarrow{\boldsymbol{h}}_2, \dots \overrightarrow{\boldsymbol{h}}_N]$ are generated for representing the input word sequence $x_1, x_2, \dots, x_N$ in the right-to-left and left-to-right directions, respectively, where for each word $x_i$, 
\begin{align*}
\overleftarrow{\boldsymbol{h}}_i &= \textrm{LSTM}(\overleftarrow{\boldsymbol{h}}_{i+1}, \boldsymbol{e}_{x_i}) \\
\overrightarrow{\boldsymbol{h}}_i &= \textrm{LSTM}(\overrightarrow{\boldsymbol{h}}_{i-1}, \boldsymbol{e}_{x_i})
\end{align*}
$\boldsymbol{e}_{x_i}$ is the embedding of word $x_i$.

\subsection{Attention-based decoder}
\label{sec:base_dec}

The decoder yields a word sequence in the target language  $y_1, y_2, \dots, y_M$ by calculating a sequence of hidden states $\boldsymbol{s}_1, \boldsymbol{s}_2 \dots, \boldsymbol{s}_M$ recurrently.
We use an attention-based LSTM decoder \cite{bahdanau2015neural}, where the attention memory ($\boldsymbol{H}$) is the concatenation of the attention vectors among all source words. 
Each attention vector $\boldsymbol{h}_i$ is the concatenation of the encoder states of an input token in both directions ($\overleftarrow{\boldsymbol{h}}_i$ and $\overrightarrow{\boldsymbol{h}}_i$):
\begin{align*}
\boldsymbol{h}_i &= [\overleftarrow{\boldsymbol{h}}_i; \overrightarrow{\boldsymbol{h}}_i] \\
\boldsymbol{H} &= [\boldsymbol{h}_1; \boldsymbol{h}_2; \dots; \boldsymbol{h}_N]
\end{align*}
$N$ is the number of source words.

While generating the $m$-th word, the decoder considers four factors: 
(1) the attention memory $\boldsymbol{H}$; 
(2) the previous hidden state of the LSTM model $\boldsymbol{s}_{m-1}$; 
(3) the embedding of the current input (previously generated word) $\boldsymbol{e}_{y_m}$; 
and (4) the previous context vector $\boldsymbol{\zeta}_{m-1}$ from attention memory $\boldsymbol{H}$. 
When $m=1$, we initialize $\boldsymbol{\zeta}_{0}$ as a zero vector, set $\boldsymbol{e}_{y_1}$ to the embedding of sentence start token ``$<$s$>$'', and calculate $\boldsymbol{s}_{0}$ from the last step of the encoder states via a dense layer:
\[
\boldsymbol{s}_{0} = \boldsymbol{W}_1 [\overleftarrow{\boldsymbol{h}}_0; \overrightarrow{\boldsymbol{h}}_N] + \boldsymbol{b}_1
\]
where $\boldsymbol{W}_1$ and $\boldsymbol{b}_1$ are model parameters.

For each decoding step $m$, the decoder feeds the concatenation of the embedding of the current input $\boldsymbol{e}_{y_m}$ and the previous context vector $\boldsymbol{\zeta}_{m-1}$ into the LSTM model to update its hidden state:
\[
\boldsymbol{s}_m = \textrm{LSTM}(\boldsymbol{s}_{m-1}, [\boldsymbol{e}_{y_m};\boldsymbol{\zeta}_{m-1}])
\]
Then the attention probability $\alpha_{m,i}$ on the attention vector $\boldsymbol{h}_i \in \boldsymbol{H}$ for the current decode step is calculated as:
\begin{align*}
\epsilon_{m,i} &= \boldsymbol{v}_2^\intercal \tanh(\boldsymbol{W}_h \boldsymbol{h}_i + \boldsymbol{W}_s \boldsymbol{s}_m + \boldsymbol{b}_2) \\
\alpha_{m,i} &= \frac{\exp(\epsilon_{m,i})}{\sum_{j=1}^N\exp(\epsilon_{m,j})} 
\end{align*}
$\boldsymbol{W}_h$, $\boldsymbol{W}_s$, $\boldsymbol{v}_2$ and $\boldsymbol{b}_2$ are model parameters.
The new context vector $\boldsymbol{\zeta}_m$ is calculated via 
\[
\boldsymbol{\zeta}_m = \sum_{i=1}^N \alpha_{m,i} \boldsymbol{h}_{i}
\]
The output probability distribution over the target vocabulary at the current state is calculated by:
\begin{equation}
\boldsymbol{P}_{vocab} = \textrm{softmax}(\boldsymbol{V}_3[\boldsymbol{s}_m,\boldsymbol{\zeta}_m]+\boldsymbol{b}_3)\textrm{,}
\label{eq:pvocab}
\end{equation}
where $\boldsymbol{V}_3$ and $\boldsymbol{b}_3$ are learnable parameters.

\section{Incorporating AMR}
\label{sec:amr}

\begin{figure}
\centering
\includegraphics[width=0.9\linewidth]{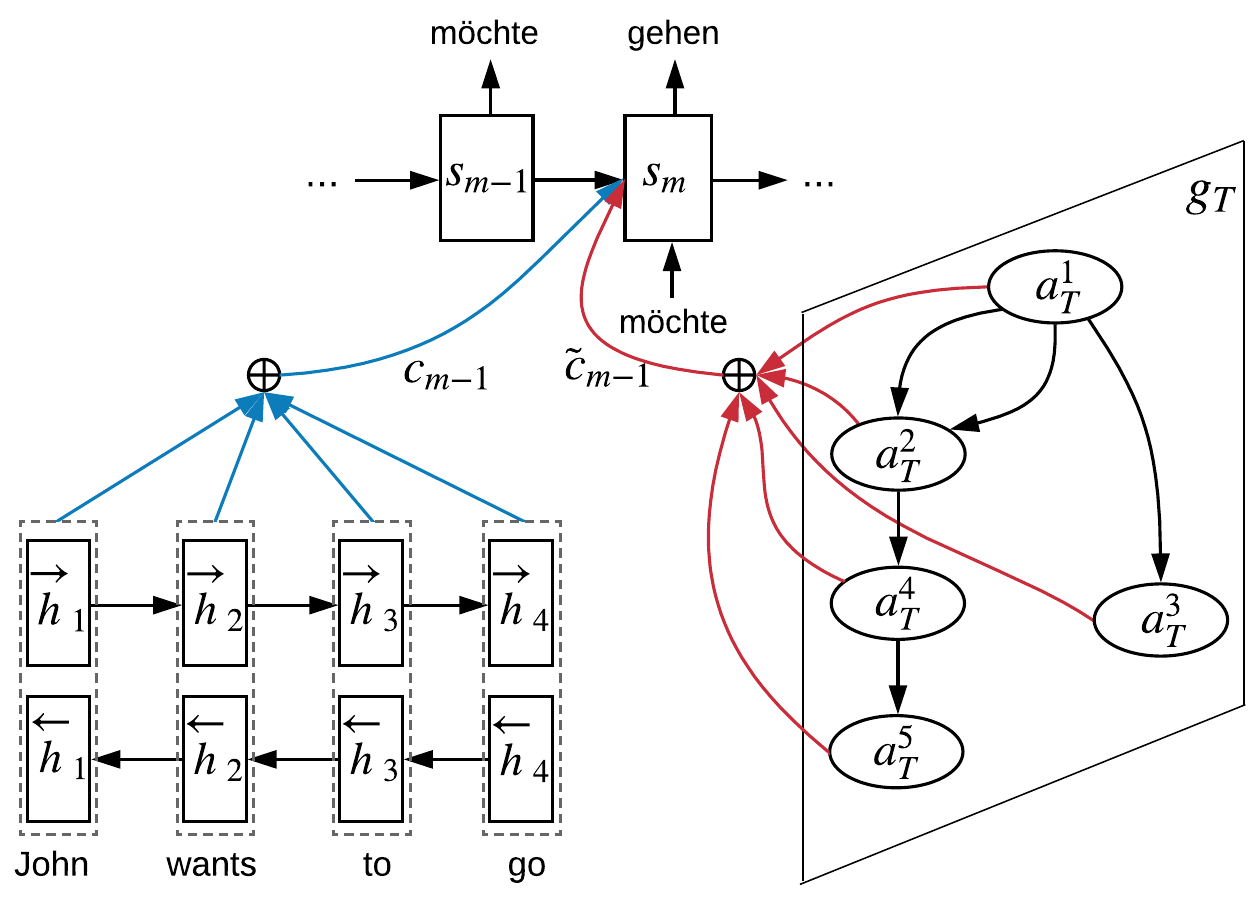}
\vspace{-1.0em}
\caption{Overall architecture of our model.}
\label{fig:dual2seq}
\vspace{-1.0em}
\end{figure}

Figure \ref{fig:dual2seq} shows the overall architecture of our model, which adopts a BiLSTM (bottom left) and our graph recurrent network (GRN)\footnote{We show the advantage of our graph encoder by comparing with another popular way for encoding AMRs in Section \ref{sec:main_res}.} (bottom right) for encoding the source sentence and AMR, respectively.
An attention-based LSTM decoder is used to generate the output sequence in the target language, with attention models over both the sequential encoder and the graph encoder.
The attention memory for the graph encoder is from the last step of the graph state transition process, which is shown in Figure \ref{fig:encoder}.

\subsection{Encoding AMR with GRN}

\begin{figure}
\centering
\includegraphics[width=0.75\linewidth]{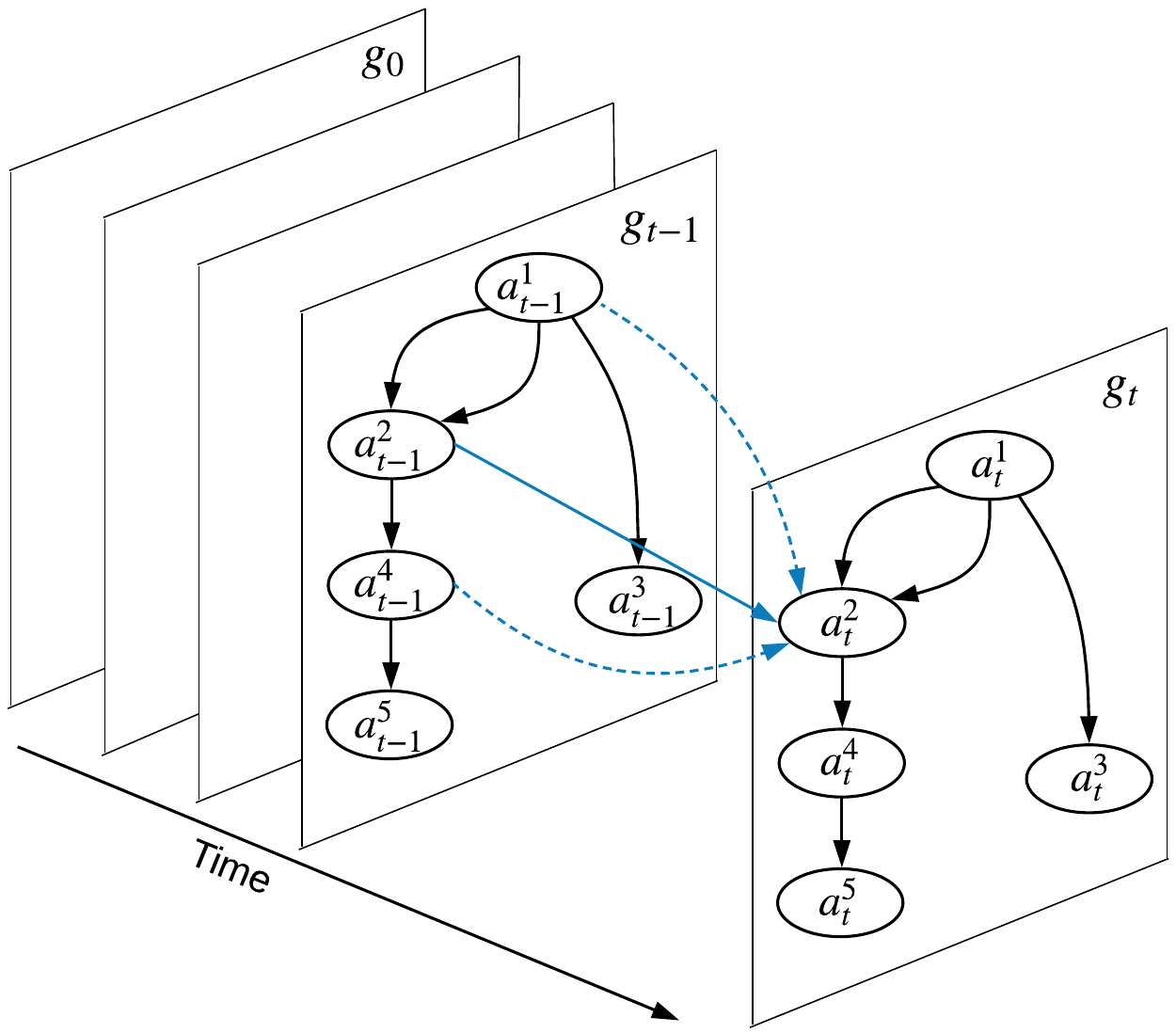}
\vspace{-0.8em}
\caption{Architecture of the graph recurrent network.}
\label{fig:encoder}
\vspace{-1.0em}
\end{figure}

Figure \ref{fig:encoder} shows the overall structure of our graph recurrent network for encoding AMR graphs, which follows \newcite{song-EtAl:acl2018}.
Formally, given an AMR graph $\boldsymbol{G}=(\boldsymbol{V}, \boldsymbol{E})$, we use a hidden state vector $\boldsymbol{a}^j$ to represent each node $v_j \in \boldsymbol{V}$. 
The state of the graph can thus be represented as:
\[
\boldsymbol{g} = \{\boldsymbol{a}^j\}|_{v_j \in \boldsymbol{V}}
\]
In order to capture non-local interaction between nodes, information exchange between nodes is executed through a sequence of state transitions, leading to a sequence of states $\boldsymbol{g}_0, \boldsymbol{g}_1, \dots, \boldsymbol{g}_T$, where $\boldsymbol{g}_t = \{\boldsymbol{a}_t^j\}|_{v_j \in \boldsymbol{V}}$, and $T$ is the number of state transitions, which is a hyperparameter.
The initial state $\boldsymbol{g}_0$ consists of a set of initial node states $\boldsymbol{a}_0^j=\boldsymbol{a}_0$, where $\boldsymbol{a}_0$ is a vector of all zeros.

A recurrent neural network is used to model the state transition process. 
In particular, the transition from $\boldsymbol{g}_{t-1}$ to $\boldsymbol{g}_t$ consists of a hidden state transition for each node (such as from $\boldsymbol{a}_{t-1}^j$ to $\boldsymbol{a}_t^j$), as shown in Figure \ref{fig:encoder}. 
At each state transition step $t$, our model conducts direct communication between a node and all nodes that are directly connected to the node. 
To avoid gradient diminishing or bursting, LSTM \cite{hochreiter1997long} is adopted, where a cell $\boldsymbol{c}_t^j$ is taken to record memory for $\boldsymbol{a}_t^j$. 
We use an input gate $\boldsymbol{i}_t^j$, an output gate $\boldsymbol{o}_t^j$ and a forget gate $\boldsymbol{f}_t^j$ to control information flow from the inputs and to the output $\boldsymbol{a}_t^j$.

The inputs include representations of edges that are connected to $v_j$, where $v_j$ can be either the source or the target of the edge.
We define each edge as a triple $(i,j,l)$, where $i$ and $j$ are indices of the source and target nodes, respectively, and $l$ is the edge label.
$\boldsymbol{x}_{i,j}^l$ is the representation of edge $(i,j,l)$, detailed in Section \ref{sec:amr_input}.
The inputs for $v_j$ are grouped into incoming and outgoing edges, before being summed up:
\begin{equation*}
\begin{split}
\boldsymbol{\phi}_j &= \sum_{(i,j,l)\in \boldsymbol{E}_{in}(j)} \boldsymbol{x}_{i,j}^l \\
\hat{\boldsymbol{\phi}}_j &= \sum_{(j,k,l)\in \boldsymbol{E}_{out}(j)} \boldsymbol{x}_{j,k}^l \\
\end{split}
\end{equation*}
where $\boldsymbol{E}_{in}(j)$ and $\boldsymbol{E}_{out}(j)$ are the sets of incoming and outgoing edges of $v_j$, respectively.

In addition to edge inputs, our model also takes the hidden states of the incoming and outgoing neighbors of each node during a state transition. 
Taking $v_j$ as an example, the states of its incoming and outgoing neighbors are summed up before being passed to the cell and gate nodes:
\begin{equation*}
\begin{split}
\boldsymbol{\psi}_j &= \sum_{(i,j,l)\in \boldsymbol{E}_{in}(j)} \boldsymbol{a}_{t-1}^{i} \\
\hat{\boldsymbol{\psi}}_j &= \sum_{(j,k,l)\in \boldsymbol{E}_{out}(j)} \boldsymbol{a}_{t-1}^{k} \textrm{,} \\
\end{split}
\end{equation*} 
Based on the above definitions of $\boldsymbol{\phi}_j$, $\hat{\boldsymbol{\phi}}_j$, $\boldsymbol{\psi}_j$ and $\hat{\boldsymbol{\psi}}_j$, the state transition from $\boldsymbol{g}_{t-1}$ to $\boldsymbol{g}_t$, as represented by $\boldsymbol{a}_t^j$, can be defined as:
\begin{equation*} \small
\begin{split}
\boldsymbol{i}_t^j &= \sigma(\boldsymbol{W}_i \boldsymbol{\phi}_j + \hat{\boldsymbol{W}}_i \hat{\boldsymbol{\phi}}_j + \boldsymbol{U}_i \boldsymbol{\psi}_j + \hat{\boldsymbol{U}}_i \hat{\boldsymbol{\psi}}_j + \boldsymbol{b}_i) \\
\boldsymbol{o}_t^j &= \sigma(\boldsymbol{W}_o \boldsymbol{\phi}_j + \hat{\boldsymbol{W}}_o \hat{\boldsymbol{\phi}}_j + \boldsymbol{U}_o \boldsymbol{\psi}_j + \hat{\boldsymbol{U}}_o \hat{\boldsymbol{\psi}}_j + \boldsymbol{b}_o) \\
\boldsymbol{f}_t^j &= \sigma(\boldsymbol{W}_f \boldsymbol{\phi}_j + \hat{\boldsymbol{W}}_f \hat{\boldsymbol{\phi}}_j + \boldsymbol{U}_f \boldsymbol{\psi}_j + \hat{\boldsymbol{U}}_f \hat{\boldsymbol{\psi}}_j + \boldsymbol{b}_f) \\
\boldsymbol{u}_t^j &= \sigma(\boldsymbol{W}_u \boldsymbol{\phi}_j + \hat{\boldsymbol{W}}_u \hat{\boldsymbol{\phi}}_j + \boldsymbol{U}_u \boldsymbol{\psi}_j + \hat{\boldsymbol{U}}_u \hat{\boldsymbol{\psi}}_j + \boldsymbol{b}_u) \\
\boldsymbol{c}_t^j &= \boldsymbol{f}_t^j \odot \boldsymbol{c}_{t-1}^j + \boldsymbol{i}_t^j \odot \boldsymbol{u}_t^j \\
\boldsymbol{a}_t^j &= \boldsymbol{o}_t^j \odot \tanh (\boldsymbol{c}_t^j) \textrm{,} \\
\end{split}
\end{equation*}
where $\boldsymbol{i}_t^j$, $\boldsymbol{o}_t^j$ and $\boldsymbol{f}_t^j$ are the input, output and forget gates mentioned earlier. 
$\boldsymbol{W}_x$, $\hat{\boldsymbol{W}}_x$, $\boldsymbol{U}_x$, $\hat{\boldsymbol{U}}_x$, $\boldsymbol{b}_x$, where $x \in \{i, o, f, u\}$, are model parameters.

With this state transition mechanism, information of each node is propagated to all its neighboring nodes after each step. 
So after several transition steps, each node state contains the information of a large context, including its ancestors, descendants, and siblings.
For the worst case where the input graph is a chain of nodes, the maximum number of steps necessary for information from one arbitrary node to reach another is equal to the size of the graph. 
We experiment with different numbers of transition steps to study the effectiveness of global encoding.

\subsubsection{Input Representation}
\label{sec:amr_input}
The edges of an AMR graph contain labels, which represent relations between the nodes they connect, and are thus important for modeling the graphs.
The representation for each edge $(i,j,l)$ is defined as: 
\begin{align*}
\boldsymbol{x}_{i,j}^l = \boldsymbol{W}_4 \Big( [\boldsymbol{e}_l; \boldsymbol{e}_{v_i}] \Big) + \boldsymbol{b}_4 \textrm{,}
\end{align*}
where $\boldsymbol{e}_l$ and $\boldsymbol{e}_i$ are the embeddings of edge label $l$ and source node $v_i$, and $\boldsymbol{W}_4$ and $\boldsymbol{b}_4$ are model parameters.

\subsection{Incorporating AMR information with a doubly-attentive decoder}

There is no one-to-one correspondence between AMR nodes and source words.
To incorporate additional knowledge from an AMR graph, an external attention model is adopted over the baseline model.
In particular, the attention memory from the AMR graph is the last graph state $\boldsymbol{g}_T = \{\boldsymbol{a}_T^j\}|_{v_j \in \boldsymbol{V}}$.
In addition, the contextual vector based on the graph state is calculated as:
\begin{align*}
\tilde{\epsilon}_{m,i} &= \tilde{\boldsymbol{v}}_2^\intercal \tanh(\boldsymbol{W}_a \boldsymbol{a}_T^i + \tilde{\boldsymbol{W}}_s \boldsymbol{s}_m + \tilde{\boldsymbol{b}}_2) \\
\tilde{\alpha}_{m,i} &= \frac{\exp(\tilde{\epsilon}_{m,i})}{\sum_{j=1}^N\exp(\tilde{\epsilon}_{m,j})} 
\end{align*}
$\boldsymbol{W}_a$, $\tilde{\boldsymbol{W}}_s$, $\tilde{\boldsymbol{v}}_2$ and $\tilde{\boldsymbol{b}}_2$ are model parameters.
The new context vector $\tilde{\boldsymbol{\zeta}}_m$ is calculated via $\sum_{i=1}^N \tilde{\alpha}_{m,i} \boldsymbol{a}_T^i$.
Finally, $\tilde{\boldsymbol{\zeta}}_m$ is incorporated into the calculation of the output probability distribution over the target vocabulary (previously defined in Equation \ref{eq:pvocab}):
\begin{equation}
\boldsymbol{P}_{vocab} = \textrm{softmax}(\boldsymbol{V}_3[\boldsymbol{s}_m,\boldsymbol{\zeta}_m,\tilde{\boldsymbol{\zeta}}_m]+\boldsymbol{b}_3)
\end{equation}

\section{Training}

Given a set of training instances $\{(\boldsymbol{X}^{(1)}, \boldsymbol{Y}^{(1)}),$ $(\boldsymbol{X}^{(2)}, \boldsymbol{Y}^{(2)}), \dots\}$,
we train our models using the cross-entropy loss over each gold-standard target sequence $\boldsymbol{Y}^{(j)}=y_1^{(j)}, y_2^{(j)}, \dots, y_M^{(j)}$:
\begin{equation*}
l = -\sum_{m=1}^M \log p(y_m^{(j)}|y_{m-1}^{(j)},\dots,y_1^{(j)},\boldsymbol{X}^{(j)};\boldsymbol{\theta})
\end{equation*}
$\boldsymbol{X}^{(j)}$ represents the inputs for the $j$th instance, which is a source sentence for our baseline, or a source sentence paired with an automatically parsed AMR graph for our model. $\boldsymbol{\theta}$ represents the model parameters.

\section{Experiments}

We empirically investigate the effectiveness of AMR for English-to-German translation.

\subsection{Setup}

\begin{table}
\centering
\begin{tabular}{lccc}
\hline
Dataset & \#Sent. & \#Tok. (EN) & \#Tok. (DE) \\
\hline
NC-v11 & 226K & 6.4M & 7.3M \\
Full & 4.17M & 109M & 118M \\
News2013 & 3000 & 84.7K & 95.6K \\
News2016 & 2999 & 88.1K & 98.8K \\
\hline
\end{tabular}
\caption{Statistics of the dataset. Numbers of tokens are after BPE processing.}
\label{tab:stat}
\end{table}

\begin{table}
\centering
\begin{tabular}{lcccc}
\hline
Dataset & EN-ori & EN & AMR & DE \\
\hline
NC-v11 & 79.8K & 8.4K & 36.6K & 8.3K \\
Full & 874K & 19.3K & 403K & 19.1K \\
\hline
\end{tabular}
\caption{Sizes of vocabularies. \emph{EN-ori} represents original English sentences without BPE.}
\label{tab:stat_vocab}
\end{table}

\subparagraph{Data}
We use the WMT16\footnote{http://www.statmt.org/wmt16/translation-task.html} English-to-German dataset, which contains around 4.5 million sentence pairs for training.
In addition, we use a subset of the full dataset (News Commentary v11 (NC-v11), containing around 243 thousand sentence pairs) for development and additional experiments.
For all experiments, we use newstest2013 and newstest2016 as the development and test sets, respectively.

To preprocess the data, the tokenizer from Moses\footnote{http://www.statmt.org/moses/} is used to tokenize both the English and German sides. 
The training sentence pairs where either side is longer than 50 words are filtered out after tokenization.
To deal with rare and compound words, byte-pair encoding (BPE)\footnote{https://github.com/rsennrich/subword-nmt} \cite{sennrich-haddow-birch:2016:P16-12} is applied to both sides.
In particular, 8000 and 16000 BPE merges are used on the News Commentary v11 subset and the full training set, respectively.
On the other hand, JAMR\footnote{https://github.com/jflanigan/jamr} \cite{flanigan-EtAl:2016:SemEval} is adopted to parse the English sentences into AMRs before BPE is applied.
The statistics of the training data and vocabularies after preprocessing are shown in Table \ref{tab:stat} and \ref{tab:stat_vocab}, respectively.
For the experiments with the full training set, we used the top 40K of the AMR vocabulary, which covers more than 99.6\% of the training set.

For our dependency-based and SRL-based baselines (which will be introduced in \textbf{Baseline systems}), we choose Stanford CoreNLP \cite{manning-EtAl:2014:P14-5} and IBM SIRE to generate dependency trees and semantic roles, respectively.
Since both dependency trees and semantic roles are based on the original English sentences without BPE, we used the top 100K frequent English words, which cover roughly 99.0\% of the training set.

\subparagraph{Hyperparameters}
We use the Adam optimizer \cite{kingma2014adam} with a learning rate of 0.0005.
The batch size is set to 128.
Between layers, we apply dropout with a probability of 0.2.
The best model is picked based on the cross-entropy loss on the development set.
For model hyperparameters, we set the graph state transition number to 10 according to development experiments.
Each node takes information from at most 6 neighbors. 
BLEU \cite{papineni-EtAl:2002:ACL}, TER \cite{snover2006study} and Meteor \cite{denkowski:lavie:meteor-wmt:2014} are used as the metrics on cased and tokenized results.

For experiments with the NC-v11 subset, both word embedding and hidden vector sizes are set to 500, and the models are trained for at most 30 epochs.
For experiments with full training set, the word embedding and hidden state sizes are set to 800, and our models are trained for at most 10 epochs.
For all systems, the word embeddings are randomly initialized and updated during training.

\begin{table*}
\centering
\begin{tabular}{l|c|c|c||c|c|c}
\multirow{2}{*}{System} & \multicolumn{3}{c||}{\textsc{NC-v11}} & \multicolumn{3}{c}{\textsc{Full}} \\
     & BLEU & TER$\downarrow$ & Meteor & BLEU & TER$\downarrow$ & Meteor \\
\hline
OpenNMT-tf & 15.1 & 0.6902 & 0.3040 & 24.3 & 0.5567 & 0.4225 \\
Transformer-tf & 17.1 & 0.6647 & 0.3578 & 25.1 & 0.5537 & 0.4344 \\
\hline
Seq2seq & 16.0 & 0.6695 & 0.3379 & 23.7 & 0.5590 & 0.4258 \\
Dual2seq-LinAMR & 17.3 & 0.6530 & 0.3612 & 24.0 & 0.5643 & 0.4246 \\
Duel2seq-SRL & 17.2 & 0.6591 & 0.3644 & 23.8 & 0.5626 &  0.4223 \\
Dual2seq-Dep & 17.8 & 0.6516 & 0.3673 & 25.0 & 0.5538 & 0.4328 \\
Dual2seq & \textbf{\phantom{*}19.2*} & \textbf{0.6305} & \textbf{0.3840} & \textbf{\phantom{*}25.5*} & \textbf{0.5480} & \textbf{0.4376} \\
\end{tabular}
\caption{\textsc{Test} performance. \emph{NC-v11} represents training only with the NC-v11 data, while \emph{Full} means using the full training data. * represents significant \cite{koehn2004statistical} result ($p < 0.01$) over \emph{Seq2seq}. $\downarrow$ indicates the lower the better.}
\label{tab:test}
\end{table*}

\begin{figure}
\includegraphics[width=0.95\linewidth]{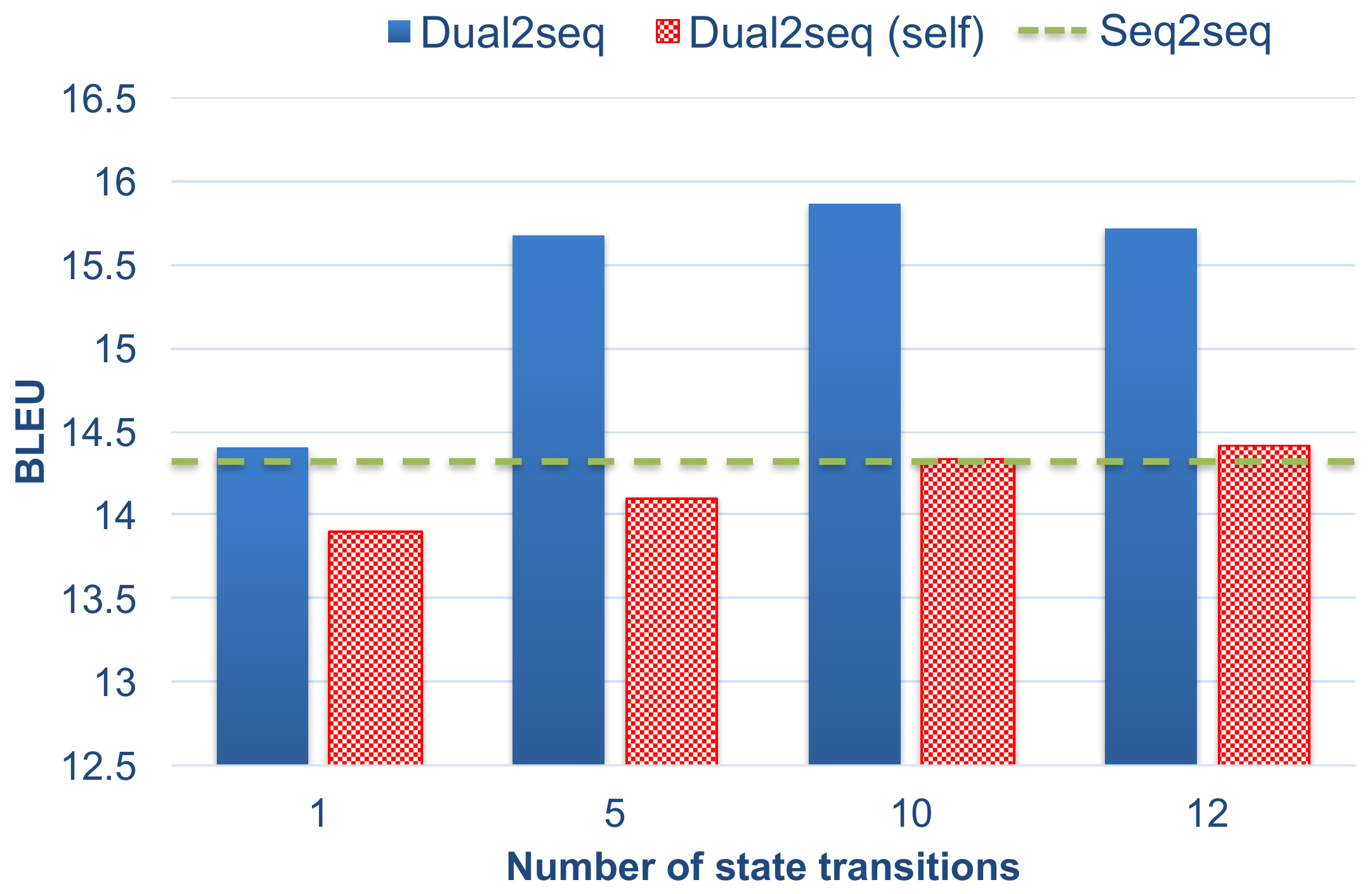}
\vspace{-0.5em}
\caption{\textsc{Dev} BLEU scores against transition steps for the graph encoders. 
The state transition is not applicable to \emph{Seq2seq}, so we draw a dashed line to represent its performance.}
\label{fig:dev_ana}
\vspace{-0.5em}
\end{figure}

\subparagraph{Baseline systems}
We compare our model with the following systems.
\emph{Seq2seq} represents our attention-based LSTM baseline ({\S \ref{sec:base}}), and \emph{Dual2seq} is our model, which takes both a sequential 
and a graph encoder and adopts a doubly-attentive decoder ({\S \ref{sec:amr}}).
To show the merit of AMR, we further contrast our model with the following baselines, all of which adopt the same doubly-attentive framework with a BiLSTM for encoding BPE-segmented source sentences: \emph{Dual2seq-LinAMR} uses another BiLSTM for encoding linearized AMRs. 
\emph{Dual2seq-Dep} and \emph{Dual2seq-SRL} adopt our graph recurrent network to encode original source sentences with dependency and semantic role annotations, respectively.
The three baselines are useful for contrasting different methods of encoding AMRs and for comparing AMRs with other popular structural information for NMT.

We also compare with Transformer \cite{NIPS2017_7181} and OpenNMT \cite{2017opennmt}, trained on the same dataset and with the same set of hyperparameters as our systems.
In particular, we compare with \emph{Transformer-tf}, one popular implementation\footnote{https://github.com/Kyubyong/transformer} of Transformer based on TensorFlow, and we choose \emph{OpenNMT-tf}, an official release\footnote{https://github.com/OpenNMT/OpenNMT-tf} of OpenNMT implemented with TensorFlow.
For a fair comparison, \emph{OpenNMT-tf} has 1 layer for both the encoder and the decoder, and \emph{Transformer-tf} has the default configuration (N=6), but with parameters being shared among different blocks.

\subsection{Development experiments}

Figure \ref{fig:dev_ana} shows the system performances as a function of the number of graph state transitions on the development set.
\emph{Dual2seq (self)} represents our dual-attentive model, but its graph encoder encodes the source sentence, which is treated as a chain graph, instead of an AMR graph.
Compared with \emph{Dual2seq}, \emph{Dual2seq (self)} has the same number of parameters, but without semantic information from AMR\@.
Due to hardware limitations, we do not perform an exhaustive search by evaluating every possible state transition number, but only transition numbers of 1, 5, 10 and 12.

Our \emph{Dual2seq} shows consistent performance improvement by increasing the transition number both from 1 to 5 (roughly +1.3 BLEU points) and from 5 to 10 (roughly 0.2 BLEU points).
The former shows greater improvement than the latter, showing that the performance starts to converge after 5 transition steps.
Further increasing transition steps from 10 to 12 gives a slight performance drop.
We set the number of state transition steps to 10 for all experiments according to these observations.

On the other hand, \emph{Dual2seq (self)} shows only small improvements
by increasing the state transition number, and it does not perform better than \emph{Seq2seq}.
Both results show that the performance gains of \emph{Dual2seq} are not due to an increased number of parameters.

\subsection{Main results}
\label{sec:main_res}

Table \ref{tab:test} shows the \textsc{test} BLEU, TER and Meteor scores of all systems trained on the small-scale \emph{News Commentary v11} subset or the large-scale full set.
\emph{Dual2seq} is consistently better than the other systems under all three metrics, showing the effectiveness of the semantic information provided by AMR\@.
Especially, \emph{Dual2seq} is better than both \emph{OpenNMT-tf} and \emph{Transformer-tf}\@.
The recurrent graph state transition of \emph{Dual2seq} is similar to Transformer in 
that it iteratively incorporates global information.
The improvement of \emph{Dual2seq} over \emph{Transformer-tf} undoubtedly comes from the use of AMRs,
 which provide complementary information to the textual inputs of the source language.

In terms of BLEU score, \emph{Dual2seq} is significantly better than \emph{Seq2seq} in both settings, which shows the effectiveness of incorporating AMR information.
In particular, the improvement is much larger under the small-scale setting (+3.2 BLEU) than that under the large-scale setting (+1.7 BLEU).
This is an evidence that structural and coarse-grained semantic information encoded in AMRs can be more helpful when training data are limited.

When trained on the NC-v11 subset, the gap between \emph{Seq2seq} and \emph{Dual2seq} under Meteor (around 5 points) is greater than that under BLEU (around 3 points).
Since Meteor gives partial credit to outputs that are synonyms to the reference or share identical stems, one possible explanation is that the structural information within AMRs helps to better translate the concepts from the source language, which may be synonyms or paronyms of reference words.

As shown in the second group of Table \ref{tab:test}, we further compare our model with other methods of leveraging syntactic or semantic information. 
\emph{Dual2seq-LinAMR} shows much worse performance than our model and only slightly outperforms the \emph{Seq2seq} baseline.
Both results show that simply taking advantage of the AMR concepts without their relations does not help very much.
One reason may be that AMR concepts, such as {\em John} and {\em Mary}, also appear in the textual input, and thus are also encoded by the other (sequential) encoder.\footnote{AMRs can contain multi-word concepts, such as {\em New York City}, but they are in the textual input.}
The gap between \emph{Dual2seq} and \emph{Dual2seq-LinAMR} comes from modeling the relations between concepts, which can be helpful for deciding target word order by enhancing the relations in source sentences.
We conclude that properly encoding AMRs is necessary to make them useful.

Encoding dependency trees instead of AMRs, \emph{Dual2seq-Dep} shows a larger performance gap with our model (17.8 vs 19.2) on small-scale training data than on large-scale training data (25.0 vs 25.5).
It is likely because AMRs are more useful on alleviating data sparsity than dependency trees, since words are lemmatized into unified concepts when parsing sentences into AMRs.
For modeling long-range dependencies, AMRs have one crucial advantage over dependency trees by modeling concept-concept relations more directly.
It is because AMRs drop function words, thus the distances between concepts are generally closer in AMRs than in dependency trees.
Finally, \emph{Dual2seq-SRL} is less effective than our model, because
the annotations labeled by SRL are a subset of AMRs.


We outperform \newcite{diego-EtAl:2018:PAPERS} on the same datasets,
although our systems vary in a number of respects.
When trained on the \emph{NC-v11} data, they show BLEU scores of 14.9 only with their BiLSTM baseline, 16.1 using additional dependency information, 15.6 using additional semantic roles and 15.8 taking both as additional knowledge.
Using \emph{Full} as the training data, the scores become 23.3, 23.9, 24.5 and 24.9, respectively.
In addition to the different semantic representation being used (AMR vs SRL), \newcite{diego-EtAl:2018:PAPERS} laid graph convolutional network (GCN) \cite{kipf2017semi} layers on top of a bidirectional LSTM (BiLSTM) layer, and then concatenated layer outputs as the attention memory. 
GCN layers encode the semantic role information, while BiLSTM layers encode the input sentence in the source language, and the concatenated hidden states of both layers contain information from both semantic role and source sentence.
For incorporating AMR, since there is no one-to-one word-to-node correspondence between a sentence and the corresponding AMR graph, we adopt separate attention models.
Our BLEU scores are higher than theirs, but we cannot conclude that the advantage primarily comes from AMR.

\subsection{Analysis}
\label{sec:analyze}

\begin{table}[t]
\centering
\begin{tabular}{l|c}
AMR Anno. & BLEU  \\
\hline
Automatic & 16.8 \\
Gold & \textbf{\phantom{*}17.5*} \\
\end{tabular}
\caption{BLEU scores of \emph{Dual2seq} on the \emph{little prince} data, when gold or automatic AMRs are available.}
\label{tab:amr_accu_ana}
\end{table}

\subparagraph{Influence of AMR parsing accuracy}
To analyze the influence of AMR parsing on our model performance, we further evaluate on a test set where the gold AMRs for the English side are available. 
In particular, we choose the \emph{Little Prince} corpus, which contains 1562 sentences with gold AMR annotations.\footnote{https://amr.isi.edu/download.html}
Since there are no parallel German sentences, we take a German-version \emph{Little Prince} novel, and then perform manual sentence alignment. 
Taking the whole \emph{Little Prince} corpus as the test set, we measure the influence of AMR parsing accuracy by evaluating on the test set when gold or automatically-parsed AMRs are available.
The automatic AMRs are generated by parsing the English sentences with JAMR.

\begin{figure}
\centering
\includegraphics[width=0.95\linewidth]{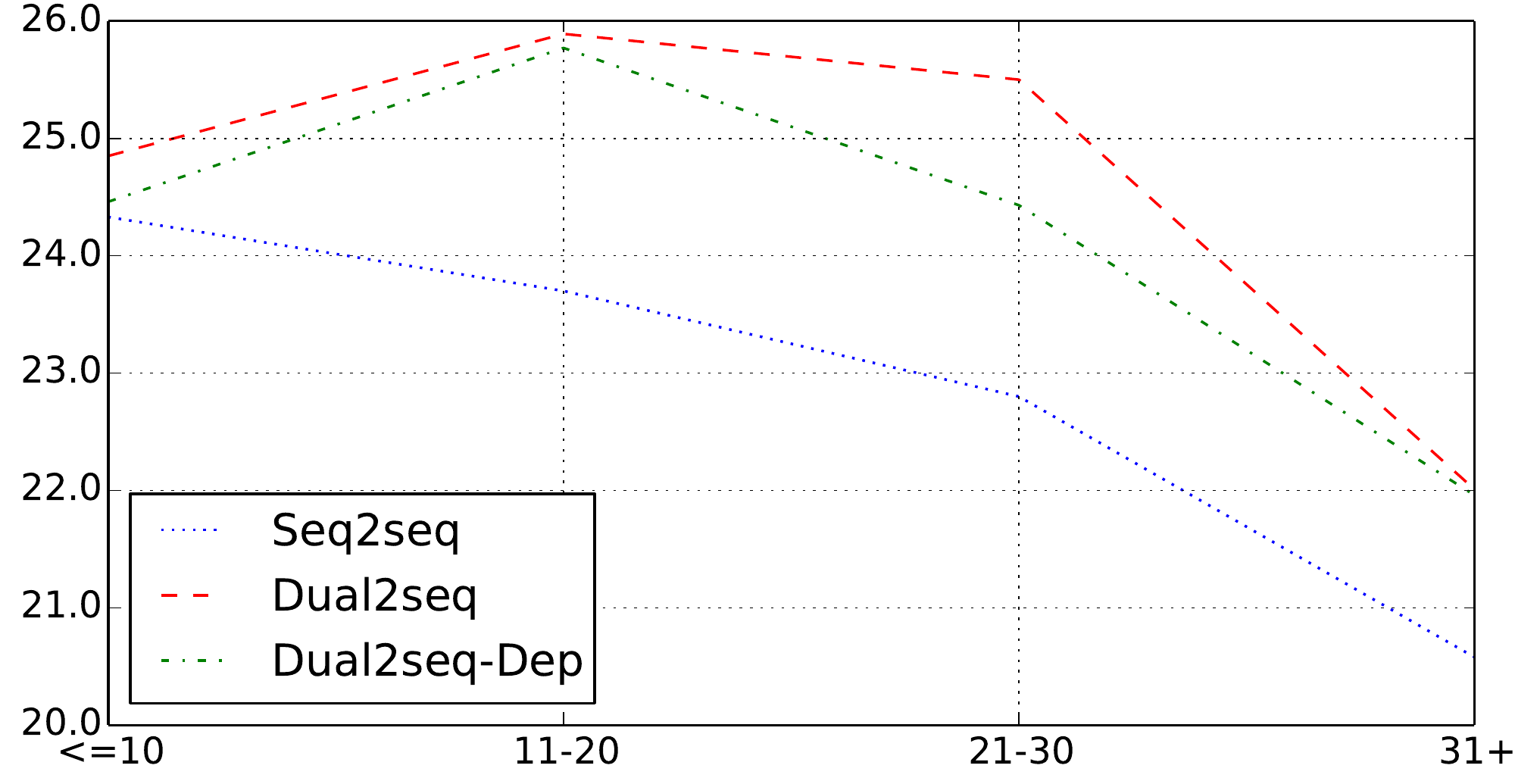}
\vspace{-1.0em}
\caption{Test BLEU score of various sentence lengths}
\label{fig:category}
\end{figure}

\begin{figure*}[t]
{\small
\begin{tabular}{p{6.25in}}
\hline
\textbf{AMR}: (s2 / say-01
       :ARG0 (p3 / person
             :ARG1-of (h / have-rel-role-91
                   :ARG0 (p / person
                         :ARG1-of (m2 / meet-03
                               :ARG0 (t / they)
                               :ARG2 15)
                         :mod (m / mutual))
                   :ARG2 (f / friend))
             :name (n2 / name
                   :op1 "Carla"
                   :op2 "Hairston"))
       :ARG1 (a / and
             :op1 (p2 / person
                   :name (n / name
                         :op1 "Lamb")))
       :ARG2 (s / she)
       :time 20) \\
\textbf{Src}: Carla Hairston said she was 15 and Lamb was 20 when they met through mutual friends .\\
\textbf{Ref}: Carla Hairston sagte , sie war 15 und Lamm war 20 , als sie sich durch gemeinsame Freunde trafen .\\
\textbf{Dual2seq}: Carla Hairston sagte , sie war 15 und Lamm war 20 , als sie sich durch gegenseitige Freunde trafen .\\
\textbf{Seq2seq}: Carla Hirston sagte , sie sei 15 und Lamb 20 , als sie durch gegenseitige Freunde trafen .\\
\hline
\textbf{AMR}: (s / say-01
       :ARG0 (m / media
             :ARG1-of (l / local-02))
       :ARG1 (c2 / come-01
             :ARG1 (v / vehicle
                   :mod (p / police))
             :manner (c3 / constant)
             :path (a / across
                   :op1 (r / refugee
                         :mod (n2 / new)))
             :time (s2 / since
                   :op1 (t3 / then))
             :topic (t / thing
                   :name (n / name
                         :op1 (c / Croatian)
                         :op2 (t2 / Tavarnik))))) \\
\textbf{Src}: Since then , according to local media , police vehicles are constantly coming across new refugees in Croatian Tavarnik .\\
\textbf{Ref}: Laut lokalen Medien treffen seitdem im kroatischen Tovarnik ständig Polizeifahrzeuge mit neuen Flüchtlingen ein .\\
\textbf{Dual2seq}: Seither kommen die Polizeifahrzeuge nach den örtlichen Medien ständig über neue Flüchtlinge in Kroatische Tavarnik .\\
\textbf{Seq2seq}: Seitdem sind die Polizeiautos nach den lokalen Medien ständig neue Flüchtlinge in Kroatien Tavarnik .\\
\hline
\textbf{AMR}: (b2 / breed-01
       :ARG0 (p2 / person
             :ARG0-of (h / have-org-role-91
                   :ARG2 (s3 / scientist)))
       :ARG1 (w2 / worm)
       :ARG2 (s2 / system
             :ARG1-of (c / control-01
                   :ARG0 (b / burst-01
                         :ARG1 (w / wave
                               :mod (s / sound)))
                   :ARG1-of (p / possible-01))
             :ARG1-of (n / nervous-01)
             :mod (m / modify-01
                   :ARG1 (g / genetics)))) \\
\textbf{Src}: Scientists have bred worms with genetically modified nervous systems that can be controlled by bursts of sound waves .\\
\textbf{Ref}: Wissenschaftler haben Würmer mit genetisch veränderten Nervensystemen gezüchtet , die von Ausbrüchen von Schallwellen gesteuert werden können .\\
\textbf{Dual2seq}: Die Wissenschaftler haben die Würmer mit genetisch veränderten Nervensystemen gezüchtet , die durch Verbrennungen von Schallwellen kontrolliert werden können .\\
\textbf{Seq2seq}: Wissenschaftler haben sich mit genetisch modifiziertem Nervensystem gezüchtet , die durch Verbrennungen von Klangwellen gesteuert werden können .\\
\hline
\end{tabular}}
\caption{Sample system outputs}\label{fig:ex}
\end{figure*}

Table \ref{tab:amr_accu_ana} shows the BLEU scores of our \emph{Dual2seq} model taking gold or automatic AMRs as inputs.
Not listed in Table \ref{tab:amr_accu_ana}, \emph{Seq2seq} achieves a BLEU score of 15.6, which is 1.2 BLEU points lower than using automatic AMR information.
The improvement from automatic AMR to gold AMR (+0.7 BLEU) is significant, which shows that the translation quality of our model can be further improved with an increase of AMR parsing accuracy.
However, the BLEU score with gold AMR does not indicate the potentially best performance that our model can achieve.
The primary reason is that even though the test set is coupled with gold AMRs, the training set is not.
Trained with automatic AMRs, our model can learn to selectively trust the AMR structure.
An additional reason is the domain difference: the \emph{Little Prince} data are in the literary domain while our training data are in the news domain.
There can be a further performance gain if the accuracy of the automatic AMRs on the training set is improved.

\subparagraph{Performance based on sentence length}
We hypothesize that AMRs should be more beneficial for longer sentences:
those are likely to contain long-distance dependencies (such as discourse information and predicate-argument structures) which may not be adequately captured by linear
chain RNNs but are directly encoded in AMRs.
To test this, we partition the test data into four buckets
by length and calculate BLEU for each of them. 
Figure \ref{fig:category} shows the performances of our model along with \emph{Dual2seq-Dep} and \emph{Seq2seq}.
Our model outperforms the \emph{Seq2seq} baseline rather uniformly across all buckets, except for the first one where they are roughly equal. 
This may be surprising.
On the one hand, \emph{Seq2seq} fails to capture some dependencies for medium-length instances;
on the other hand AMR parses are more noisy for longer sentences, which prevents us from obtaining extra improvements with AMRs.

Dependency trees have been proved useful in capturing long-range dependencies.
Figure \ref{fig:category} shows that AMRs are comparatively better than dependency trees, especially on medium-length (21-30) sentences.
The reason may be that the AMRs of medium-length sentences are much more accurate than longer sentences, thus are better at capturing the relations between concepts.
On the other hand, even though dependency trees are more accurate than AMRs, they still fail to represent relations for long sentences.
It is likely because relations for longer sentences are more difficult to detect.
Another possible reason is that dependency trees do not incorporate coreferences, which AMRs consider.

\subparagraph{Human evaluation}
We further study the translation quality of predicate-argument structures by conducting a human evaluation on 100 instances from the testset.
In the evaluation, translations of both \emph{Dual2seq} and \emph{Seq2seq}, together with the source English sentence, the German reference, and an AMR are provided to a German-speaking annotator to decide which translation better captures the predicate-argument structures in the source sentence.
To avoid annotation bias, translation results of both models are swapped for some instances, and the German annotator does not know which model each translation belongs to.
The annotator either selects a ``winner'' or makes a ``tie'' decision, meaning that both results are equally good.

Out of the 100 instances, \emph{Dual2seq} wins on 46, \emph{Seq2seq} wins on 23, and there is a tie on the remaining 31.
\emph{Dual2seq} wins on almost half of the instances,
about twice as often as \emph{Seq2seq} wins, indicating that AMRs help in translating the predicate-argument structures on the source side.

\subparagraph{Case study}
The outputs of the baseline system (\emph{Seq2seq}) and our final system (\emph{Dual2seq})
are shown in Figure~\ref{fig:ex}.
In the first sentence, the AMR-based Dual2seq system correctly produces
the reflexive pronoun {\em sich} as an argument of 
the verb {\em trafen} ({\em meet}), despite the
distance between the words in the system output, and
despite the fact that the equivalent English words
{\em each other} do not appear in the system output.
This is facilitated by the argument structure in the AMR analysis.

In the second sentence, 
the AMR-based Dual2seq system produces an overly literal
translation for the English phrasal verb 
{\em come across}.  The Seq2seq
translation, however, incorrectly
states that the police vehicles {\em are} refugees.
The difficulty for the Seq2seq probably derives in part 
from the fact that {\em are} and {\em coming} are separated by
the word {\em constantly} in the input, while the main predicate
is clear in the AMR representation.

In the third sentence, the Dual2seq system correctly translates
the object of {\em breed} as {\em worms}, while the Seq2seq
translation incorrectly states that the scientists breed {\em themselves}.
Here the difficulty is likely the distance between the 
object and the verb in the German output, which causes the
Seq2seq system to lose track of the correct input position to translate.


\section{Conclusion}

We showed that AMRs can improve neural machine translation.
In particular, the structural semantic information from AMRs can be complementary to the source textual input by introducing a higher level of information abstraction.
A graph recurrent network (GRN) is leveraged to encode AMR graphs without breaking the original graph structure, and a sequential LSTM is used to encode the source input.
The decoder is a doubly-attentive LSTM, taking the encoding results of both the graph encoder and the sequential encoder as attention memories.
Experiments on a standard benchmark showed that AMRs are helpful regardless of the sentence length, and are more effective than other more popular choices, such as dependency trees and semantic roles.

\section*{Acknowledgments}
We would like to thank the action editor and the anonymous reviewers for their insightful comments.
We also thank Kai Song from Alibaba for suggestions on large-scale training, Parker Riley for comments on the draft, and Rochester's CIRC for computational resources.

\bibliography{acl2018}

\begin{thebibliography}{47}
\expandafter\ifx\csname natexlab\endcsname\relax\def\natexlab#1{#1}\fi

\bibitem[{Aharoni and Goldberg(2017)}]{aharoni-goldberg:2017:Short}
Roee Aharoni and Yoav Goldberg. 2017.
\newblock Towards string-to-tree neural machine translation.
\newblock In \emph{Proceedings of the 55th Annual Meeting of the Association
  for Computational Linguistics (ACL-17)}, pages 132--140.

\bibitem[{Artzi et~al.(2015)Artzi, Lee, and
  Zettlemoyer}]{artzi-lee-zettlemoyer:2015:EMNLP}
Yoav Artzi, Kenton Lee, and Luke Zettlemoyer. 2015.
\newblock Broad-coverage {CCG} semantic parsing with {AMR}.
\newblock In \emph{Conference on Empirical Methods in Natural Language
  Processing (EMNLP-15)}, pages 1699--1710.

\bibitem[{Bahdanau et~al.(2015)Bahdanau, Cho, and Bengio}]{bahdanau2015neural}
Dzmitry Bahdanau, Kyunghyun Cho, and Yoshua Bengio. 2015.
\newblock Neural machine translation by jointly learning to align and
  translate.
\newblock In \emph{International Conference on Learning Representations
  (ICLR)}.

\bibitem[{Baker et~al.(2012)Baker, Bloodgood, Dorr, Callison-Burch, Filardo,
  Piatko, Levin, and Miller}]{baker2012modality}
Kathryn Baker, Michael Bloodgood, Bonnie~J Dorr, Chris Callison-Burch,
  Nathaniel~W Filardo, Christine Piatko, Lori Levin, and Scott Miller. 2012.
\newblock Modality and negation in {SIMT} use of modality and negation in
  semantically-informed syntactic {MT}.
\newblock \emph{Computational Linguistics}, 38(2):411--438.

\bibitem[{Banarescu et~al.(2013)Banarescu, Bonial, Cai, Georgescu, Griffitt,
  Hermjakob, Knight, Koehn, Palmer, and
  Schneider}]{banarescu-EtAl:2013:LAW7-ID}
Laura Banarescu, Claire Bonial, Shu Cai, Madalina Georgescu, Kira Griffitt, Ulf
  Hermjakob, Kevin Knight, Philipp Koehn, Martha Palmer, and Nathan Schneider.
  2013.
\newblock Abstract meaning representation for sembanking.
\newblock In \emph{Proceedings of the 7th Linguistic Annotation Workshop and
  Interoperability with Discourse}, pages 178--186.

\bibitem[{Bastings et~al.(2017)Bastings, Titov, Aziz, Marcheggiani, and
  Simaan}]{bastings-EtAl:2017:EMNLP2017}
Joost Bastings, Ivan Titov, Wilker Aziz, Diego Marcheggiani, and Khalil Simaan.
  2017.
\newblock Graph convolutional encoders for syntax-aware neural machine
  translation.
\newblock In \emph{Conference on Empirical Methods in Natural Language
  Processing (EMNLP-17)}, pages 1957--1967.

\bibitem[{Beck et~al.(2018)Beck, Haffari, and Cohn}]{P18-1026}
Daniel Beck, Gholamreza Haffari, and Trevor Cohn. 2018.
\newblock Graph-to-sequence learning using gated graph neural networks.
\newblock In \emph{Proceedings of the 56th Annual Meeting of the Association
  for Computational Linguistics (ACL-18)}, pages 273--283.

\bibitem[{Buys and Blunsom(2017)}]{buys-blunsom:2017:Long}
Jan Buys and Phil Blunsom. 2017.
\newblock Robust incremental neural semantic graph parsing.
\newblock In \emph{Proceedings of the 55th Annual Meeting of the Association
  for Computational Linguistics (ACL-17)}, pages 1215--1226.

\bibitem[{Chen et~al.(2017)Chen, Huang, Chiang, and
  Chen}]{chen-EtAl:2017:Long6}
Huadong Chen, Shujian Huang, David Chiang, and Jiajun Chen. 2017.
\newblock Improved neural machine translation with a syntax-aware encoder and
  decoder.
\newblock In \emph{Proceedings of the 55th Annual Meeting of the Association
  for Computational Linguistics (ACL-17)}, pages 1936--1945.

\bibitem[{Chen et~al.(2018)Chen, Wang, Utiyama, Sumita, and
  Zhao}]{chen2017syntax}
Kehai Chen, Rui Wang, Masao Utiyama, Eiichiro Sumita, and Tiejun Zhao. 2018.
\newblock Syntax-directed attention for neural machine translation.
\newblock In \emph{Proceedings of the National Conference on Artificial
  Intelligence (AAAI-18)}.

\bibitem[{Cho et~al.(2014)Cho, van Merrienboer, Gulcehre, Bahdanau, Bougares,
  Schwenk, and Bengio}]{cho-EtAl:2014:EMNLP2014}
Kyunghyun Cho, Bart van Merrienboer, Caglar Gulcehre, Dzmitry Bahdanau, Fethi
  Bougares, Holger Schwenk, and Yoshua Bengio. 2014.
\newblock Learning phrase representations using {RNN} encoder--decoder for
  statistical machine translation.
\newblock In \emph{Conference on Empirical Methods in Natural Language
  Processing (EMNLP-14)}, pages 1724--1734.

\bibitem[{Denkowski and Lavie(2014)}]{denkowski:lavie:meteor-wmt:2014}
Michael Denkowski and Alon Lavie. 2014.
\newblock Meteor universal: Language specific translation evaluation for any
  target language.
\newblock In \emph{Proceedings of the Ninth Workshop on Statistical Machine
  Translation}, pages 376--380.

\bibitem[{Flanigan et~al.(2016)Flanigan, Dyer, Smith, and
  Carbonell}]{flanigan-EtAl:2016:SemEval}
Jeffrey Flanigan, Chris Dyer, Noah~A. Smith, and Jaime Carbonell. 2016.
\newblock {CMU} at {SemEval-2016} {Task} 8: Graph-based {AMR} parsing with
  infinite ramp loss.
\newblock In \emph{Proceedings of the 10th International Workshop on Semantic
  Evaluation (SemEval-2016)}, pages 1202--1206.

\bibitem[{Flanigan et~al.(2014)Flanigan, Thomson, Carbonell, Dyer, and
  Smith}]{flanigan-EtAl:2014:P14-1}
Jeffrey Flanigan, Sam Thomson, Jaime Carbonell, Chris Dyer, and Noah~A. Smith.
  2014.
\newblock A discriminative graph-based parser for the abstract meaning
  representation.
\newblock In \emph{Proceedings of the 52nd Annual Meeting of the Association
  for Computational Linguistics (ACL-14)}, pages 1426--1436.

\bibitem[{Groschwitz et~al.(2018)Groschwitz, Lindemann, Fowlie, Johnson, and
  Koller}]{P18-1170}
Jonas Groschwitz, Matthias Lindemann, Meaghan Fowlie, Mark Johnson, and
  Alexander Koller. 2018.
\newblock {AMR} dependency parsing with a typed semantic algebra.
\newblock In \emph{Proceedings of the 56th Annual Meeting of the Association
  for Computational Linguistics (ACL-18)}, pages 1831--1841.

\bibitem[{Guo and Lu(2018)}]{D18-1198}
Zhijiang Guo and Wei Lu. 2018.
\newblock Better transition-based {AMR} parsing with a refined search space.
\newblock In \emph{Proceedings of the 56th Annual Meeting of the Association
  for Computational Linguistics (ACL-18)}, pages 1712--1722.

\bibitem[{Hochreiter and Schmidhuber(1997)}]{hochreiter1997long}
Sepp Hochreiter and J{\"u}rgen Schmidhuber. 1997.
\newblock Long short-term memory.
\newblock \emph{Neural computation}, 9(8):1735--1780.

\bibitem[{Isabelle et~al.(2017)Isabelle, Cherry, and
  Foster}]{isabelle-cherry-foster:2017:EMNLP2017}
Pierre Isabelle, Colin Cherry, and George Foster. 2017.
\newblock A challenge set approach to evaluating machine translation.
\newblock In \emph{Conference on Empirical Methods in Natural Language
  Processing (EMNLP-17)}, pages 2486--2496.

\bibitem[{Jones et~al.(2012)Jones, Andreas, Bauer, Hermann, and
  Knight}]{jones-EtAl:2012:PAPERS}
Bevan Jones, Jacob Andreas, Daniel Bauer, Karl~Moritz Hermann, and Kevin
  Knight. 2012.
\newblock Semantics-based machine translation with hyperedge replacement
  grammars.
\newblock In \emph{Proceedings of the International Conference on Computational
  Linguistics (COLING-12)}, pages 1359--1376.

\bibitem[{Kingma and Ba(2014)}]{kingma2014adam}
Diederik Kingma and Jimmy Ba. 2014.
\newblock Adam: A method for stochastic optimization.
\newblock \emph{arXiv preprint arXiv:1412.6980}.

\bibitem[{Kipf and Welling(2017)}]{kipf2017semi}
Thomas~N. Kipf and Max Welling. 2017.
\newblock Semi-supervised classification with graph convolutional networks.
\newblock In \emph{International Conference on Learning Representations
  ({ICLR})}.

\bibitem[{Klein et~al.(2017)Klein, Kim, Deng, Senellart, and
  Rush}]{2017opennmt}
Guillaume Klein, Yoon Kim, Yuntian Deng, Jean Senellart, and Alexander~M. Rush.
  2017.
\newblock {{OpenNMT}: Open-Source Toolkit for Neural Machine Translation}.
\newblock \emph{arXiv preprint arXiv:1701.02810}.

\bibitem[{Koehn(2004)}]{koehn2004statistical}
Philipp Koehn. 2004.
\newblock Statistical significance tests for machine translation evaluation.
\newblock In \emph{Conference on Empirical Methods in Natural Language
  Processing (EMNLP-04)}, pages 388--395.

\bibitem[{Konstas et~al.(2017)Konstas, Iyer, Yatskar, Choi, and
  Zettlemoyer}]{konstas-EtAl:2017:Long}
Ioannis Konstas, Srinivasan Iyer, Mark Yatskar, Yejin Choi, and Luke
  Zettlemoyer. 2017.
\newblock Neural {AMR}: Sequence-to-sequence models for parsing and generation.
\newblock In \emph{Proceedings of the 55th Annual Meeting of the Association
  for Computational Linguistics (ACL-17)}, pages 146--157.

\bibitem[{Li et~al.(2017)Li, Xiong, Tu, Zhu, Zhang, and
  Zhou}]{li-EtAl:2017:Long}
Junhui Li, Deyi Xiong, Zhaopeng Tu, Muhua Zhu, Min Zhang, and Guodong Zhou.
  2017.
\newblock Modeling source syntax for neural machine translation.
\newblock In \emph{Proceedings of the 55th Annual Meeting of the Association
  for Computational Linguistics (ACL-17)}, pages 688--697.

\bibitem[{Li et~al.(2015{\natexlab{a}})Li, Nguyen, Cao, and
  Grishman}]{li-EtAl:2015:CNewsStory}
Xiang Li, Thien~Huu Nguyen, Kai Cao, and Ralph Grishman. 2015{\natexlab{a}}.
\newblock Improving event detection with abstract meaning representation.
\newblock In \emph{Proceedings of the First Workshop on Computing News
  Storylines}, pages 11--15.

\bibitem[{Li et~al.(2015{\natexlab{b}})Li, Tarlow, Brockschmidt, and
  Zemel}]{li2015gated}
Yujia Li, Daniel Tarlow, Marc Brockschmidt, and Richard Zemel.
  2015{\natexlab{b}}.
\newblock Gated graph sequence neural networks.
\newblock \emph{arXiv preprint arXiv:1511.05493}.

\bibitem[{Liu and Gildea(2010)}]{liu-gildea:2010:PAPERS}
Ding Liu and Daniel Gildea. 2010.
\newblock Semantic role features for machine translation.
\newblock In \emph{Proceedings of the 23rd International Conference on
  Computational Linguistics (COLING-10)}, pages 716--724.

\bibitem[{Lyu and Titov(2018)}]{P18-1037}
Chunchuan Lyu and Ivan Titov. 2018.
\newblock {AMR} parsing as graph prediction with latent alignment.
\newblock In \emph{Proceedings of the 56th Annual Meeting of the Association
  for Computational Linguistics (ACL-18)}, pages 397--407.

\bibitem[{Manning et~al.(2014)Manning, Surdeanu, Bauer, Finkel, Bethard, and
  McClosky}]{manning-EtAl:2014:P14-5}
Christopher~D. Manning, Mihai Surdeanu, John Bauer, Jenny Finkel, Steven~J.
  Bethard, and David McClosky. 2014.
\newblock The {Stanford} {CoreNLP} natural language processing toolkit.
\newblock In \emph{Association for Computational Linguistics (ACL) System
  Demonstrations}, pages 55--60.

\bibitem[{Marcheggiani et~al.(2018)Marcheggiani, Bastings, and
  Titov}]{diego-EtAl:2018:PAPERS}
Diego Marcheggiani, Joost Bastings, and Ivan Titov. 2018.
\newblock Exploiting semantics in neural machine translation with graph
  convolutional networks.
\newblock In \emph{Proceedings of the 2018 Meeting of the North American
  chapter of the Association for Computational Linguistics (NAACL-18)}, pages
  486--492.

\bibitem[{Mitra and Baral(2015)}]{mitra2015addressing}
Arindam Mitra and Chitta Baral. 2015.
\newblock Locationing a question answering challenge by combining statistical
  methods with inductive rule learning and reasoning.
\newblock In \emph{Proceedings of the National Conference on Artificial
  Intelligence (AAAI-16)}.

\bibitem[{Papineni et~al.(2002)Papineni, Roukos, Ward, and
  Zhu}]{papineni-EtAl:2002:ACL}
Kishore Papineni, Salim Roukos, Todd Ward, and Wei-Jing Zhu. 2002.
\newblock {BLEU}: a method for automatic evaluation of machine translation.
\newblock In \emph{Proceedings of the 40th Annual Meeting of the Association
  for Computational Linguistics (ACL-02)}, pages 311--318.

\bibitem[{Peng et~al.(2015)Peng, Song, and Gildea}]{K15-1004}
Xiaochang Peng, Linfeng Song, and Daniel Gildea. 2015.
\newblock A synchronous hyperedge replacement grammar based approach for {AMR}
  parsing.
\newblock In \emph{Proceedings of the Nineteenth Conference on Computational
  Natural Language Learning}, pages 32--41.

\bibitem[{Peng et~al.(2018)Peng, Song, Gildea, and Satta}]{peng:2018:ACL}
Xiaochang Peng, Linfeng Song, Daniel Gildea, and Giorgio Satta. 2018.
\newblock Sequence-to-sequence models for cache transition systems.
\newblock In \emph{Proceedings of the 56th Annual Meeting of the Association
  for Computational Linguistics (ACL-18)}, pages 1842--1852.

\bibitem[{Pust et~al.(2015)Pust, Hermjakob, Knight, Marcu, and
  May}]{pust-EtAl:2015:EMNLP}
Michael Pust, Ulf Hermjakob, Kevin Knight, Daniel Marcu, and Jonathan May.
  2015.
\newblock Parsing english into abstract meaning representation using
  syntax-based machine translation.
\newblock In \emph{Conference on Empirical Methods in Natural Language
  Processing (EMNLP-15)}, pages 1143--1154.

\bibitem[{Sennrich et~al.(2016)Sennrich, Haddow, and
  Birch}]{sennrich-haddow-birch:2016:P16-12}
Rico Sennrich, Barry Haddow, and Alexandra Birch. 2016.
\newblock Neural machine translation of rare words with subword units.
\newblock In \emph{Proceedings of the 54th Annual Meeting of the Association
  for Computational Linguistics (ACL-16)}, pages 1715--1725.

\bibitem[{Snover et~al.(2006)Snover, Dorr, Schwartz, Micciulla, and
  Makhoul}]{snover2006study}
Matthew Snover, Bonnie Dorr, Richard Schwartz, Linnea Micciulla, and John
  Makhoul. 2006.
\newblock A study of translation edit rate with targeted human annotation.
\newblock In \emph{Proceedings of Association for Machine Translation in the
  Americas}, pages 223--231.

\bibitem[{Song et~al.(2018)Song, Zhang, Wang, and Gildea}]{song-EtAl:acl2018}
Linfeng Song, Yue Zhang, Zhiguo Wang, and Daniel Gildea. 2018.
\newblock A graph-to-sequence model for {AMR}-to-text generation.
\newblock In \emph{Proceedings of the 56th Annual Meeting of the Association
  for Computational Linguistics (ACL-18)}, pages 1842--1852.

\bibitem[{Stahlberg et~al.(2016)Stahlberg, Hasler, Waite, and
  Byrne}]{stahlberg-EtAl:2016:P16-2}
Felix Stahlberg, Eva Hasler, Aurelien Waite, and Bill Byrne. 2016.
\newblock Syntactically guided neural machine translation.
\newblock In \emph{Proceedings of the 54th Annual Meeting of the Association
  for Computational Linguistics (ACL-16)}, pages 299--305.

\bibitem[{Takase et~al.(2016)Takase, Suzuki, Okazaki, Hirao, and
  Nagata}]{takase-EtAl:2016:EMNLP2016}
Sho Takase, Jun Suzuki, Naoaki Okazaki, Tsutomu Hirao, and Masaaki Nagata.
  2016.
\newblock Neural headline generation on abstract meaning representation.
\newblock In \emph{Conference on Empirical Methods in Natural Language
  Processing (EMNLP-16)}, pages 1054--1059.

\bibitem[{Vaswani et~al.(2017)Vaswani, Shazeer, Parmar, Uszkoreit, Jones,
  Gomez, Kaiser, and Polosukhin}]{NIPS2017_7181}
Ashish Vaswani, Noam Shazeer, Niki Parmar, Jakob Uszkoreit, Llion Jones,
  Aidan~N Gomez, {\L}ukasz Kaiser, and Illia Polosukhin. 2017.
\newblock Attention is all you need.
\newblock In \emph{Advances in Neural Information Processing Systems 30}, pages
  5998--6008.

\bibitem[{Wang and Xue(2017)}]{wang-xue:2017:EMNLP2017}
Chuan Wang and Nianwen Xue. 2017.
\newblock Getting the most out of {AMR} parsing.
\newblock In \emph{Conference on Empirical Methods in Natural Language
  Processing (EMNLP-17)}, pages 1257--1268.

\bibitem[{Wong and Mooney(2006)}]{wong-mooney:2006:HLT-NAACL06-Main}
Yuk~Wah Wong and Raymond Mooney. 2006.
\newblock Learning for semantic parsing with statistical machine translation.
\newblock In \emph{Proceedings of the 2006 Meeting of the North American
  chapter of the Association for Computational Linguistics (NAACL-06)}, pages
  439--446.

\bibitem[{Wu and Fung(2009)}]{wu-fung:2009:NAACLHLT09-Short}
Dekai Wu and Pascale Fung. 2009.
\newblock Semantic roles for {SMT}: A hybrid two-pass model.
\newblock In \emph{Proceedings of the 2009 Meeting of the North American
  chapter of the Association for Computational Linguistics (NAACL-09)}, pages
  13--16.

\bibitem[{Wu et~al.(2017)Wu, Zhou, and Zhang}]{wu2017improved}
Shuangzhi Wu, Ming Zhou, and Dongdong Zhang. 2017.
\newblock Improved neural machine translation with source syntax.
\newblock In \emph{Proceedings of the Twenty-Sixth International Joint
  Conference on Artificial Intelligence (IJCAI-17)}, pages 4179--4185.

\bibitem[{Zhang et~al.(2018)Zhang, Liu, and Song}]{zhang-EtAl:acl2018}
Yue Zhang, Qi~Liu, and Linfeng Song. 2018.
\newblock Sentence-state {LSTM} for text representation.
\newblock In \emph{Proceedings of the 56th Annual Meeting of the Association
  for Computational Linguistics (ACL-18)}, pages 317--327.

\end{thebibliography}
\bibliographystyle{acl_natbib}


\end{document}